\documentclass[11pt, a4paper, logo, copyright,hyperref={colorlinks=true,    
    linkcolor=blue,     
    citecolor=blue,    
    urlcolor=magenta    
}
]{googledeepmind}

\usepackage[authoryear, sort&compress, round]{natbib}
\usepackage{amsmath}
\usepackage{amssymb}
\usepackage{graphicx} 
\usepackage{ulem}
\usepackage[most]{tcolorbox}
\usepackage{lipsum} 
\usepackage{xcolor} 
\usepackage{framed}

\definecolor{distillblue}{HTML}{4E79A7} 
\definecolor{distillgray}{HTML}{F2F2F2} 

\newtcolorbox{minimalleftrule}{
  enhanced, 
  breakable, 
  sharp corners,
  boxrule=0pt, 
  colback=white, 
  borderline west={.5pt}{0pt}{distillblue}, 
  left=8pt, 
  right=0pt,
  top=5pt,
  bottom=5pt,
  parbox=true, 
  before=\par\medskip\noindent,
  after=\par\medskip
}

\newtcolorbox{subtleshadeleftrule}{
  enhanced,
  breakable,
  sharp corners,
  boxrule=0pt,
  colback=distillgray!50, 
  borderline west={2pt}{0pt}{gray!60},
  left=10pt,
  right=5pt,
  top=8pt,
  bottom=8pt,
  parbox=true,
  before=\par\medskip\noindent,
  after=\par\medskip
}

\usepackage[authoryear, sort&compress, round]{natbib}

\reportnumber{001}

\renewcommand{\today}{2025-06}
\title{Because we have LLMs, we Can and Should Pursue Agentic Interpretability}

\author[1]{Been Kim, John Hewitt, Neel Nanda, Noah Fiedel, Oyvind Tafjord}
\affil[1]{Google DeepMind}

\begin{abstract}
The era of Large Language Models (LLMs) presents a new opportunity for interpretability--\textit{agentic interpretability}: a multi-turn conversation with an LLM wherein the LLM proactively assists human understanding by developing and leveraging a \textit{mental model of the user}, which in turn enables humans to develop better \textit{mental models of the LLM}. Such conversation is a new capability that traditional `inspective' interpretability methods (opening the black-box) do not use. Having a language model that aims to teach and explain---beyond just knowing how to talk---is similar to a teacher whose goal is to teach well, understanding that their success will be measured by the student's comprehension. While agentic interpretability may trade off completeness for interactivity, making it less suitable for high-stakes safety situations with potentially deceptive models, it leverages a cooperative model to discover potentially superhuman concepts that can improve humans' mental model of machines. Agentic interpretability introduces challenges, particularly in evaluation, due to what we call \textit{`human-entangled-in-the-loop'} nature (humans responses are integral part of the algorithm), making the design and evaluation difficult. We discuss possible solutions and proxy goals. As LLMs approach human parity in many tasks, agentic interpretability’s promise is to help humans learn the potentially superhuman concepts of the LLMs, rather than see us fall increasingly far from understanding them.
\end{abstract}

\begin{document}

\maketitle

\section{Introduction}

\textbf{We argue that large language models (LLMs) unlock opportunities for agentic interpretability: leveraging the model itself as a cooperative agent to help humans understand the model}.
LLM's capacity for coherent, contextual conversation enables this.
Many constituent ideas of agentic interpretability are common, e.g., multi-turn dialogues with LLMs and chains of thought~\citep{wei2022chain} that help humans to build mental models of machines via their reasoning.
We believe that the synthesis of these ideas---
machines and humans building mental model of each other---remains to be explored.

The term \textit{agentic} is overloaded and lacks universal consensus. Here, agentic signifies something more than, yet closely related to, \textit{proactive}. 
\begin{subtleshadeleftrule}
A method pursues agentic interpretability if it proactively assists human understanding in a multi-turn interactive process by developing and leveraging a mental model of the user, which in turn enables humans to develop better mental models of the LLM.
\end{subtleshadeleftrule}
This mental model (potentially implicit) is critical; to effectively aid comprehension, the model must infer the user's knowledge and confusion, akin to an effective teacher-student dynamic.
Multi-turn conversation is also essential, especially as many opaque model behaviors (e.g., why appending one seemingly meaningless token enables a jailbreak, but not when another arbitrary token is removed) may involve complex, counter-intuitive, or even superhuman knowledge.
The mental model is crucial for humans beyond doing well on the end-task; humans' mental model of the machine serves a distinct purpose of ensuring we do not fall behind in understanding increasingly powerful models.

\begin{figure*}
    \centering
    \includegraphics[width=\linewidth]{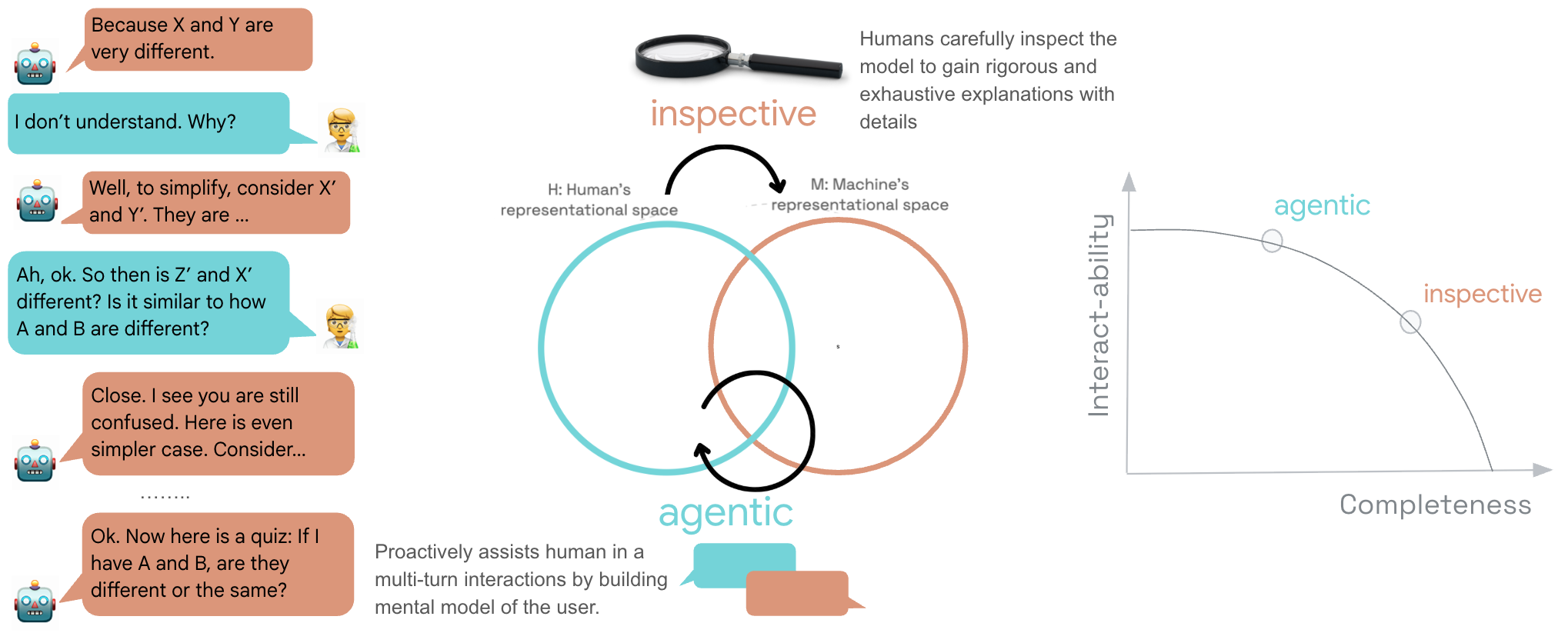}
    \caption{Agentic and inspective interpretability. Adapted from~\cite{Schut2025} with permission.}
    \label{fig:agentic_inspective}
\end{figure*}

Agentic interpretability is not the right tool for all uses of interpretability.
By focusing on interactiveness, it may sacrifice completeness, e.g. by missing some important behaviors of potentially deceptive or misaligned models.
\textit{Inspective} interpretability approaches may be a better fit for high-stakes, safety critical applications~\citep{shah2025approach, sharkey2025open,olah2020zoom}.
In contrast, agentic interpretability is particularly useful for integration of systems into society and teaching humans superhuman knowledge~\citep{Schut2025}. 

Agentic interpretability introduces evaluation challenges.
Humans are not merely in-the-loop; but interwoven in the process with machines.
We call this \textbf{human-entangled-in-the-loop}; humans' responses in the multi-turn dialogue are integral to the interpretability algorithm itself, limiting automated evaluation.
Further, variation in user backgrounds, needs and model outputs will lead to a wide range of conversational trajectories.
Finally, much knowledge might be 
beyond the capacity of an individual interacting with the system, or even superhuman in nature, making 
direct validation difficult. 
While LLMs' capability to carry out coherent conversation enables us \textbf{to some extent} to use LLMs as proxies of the human turns, end-task metrics (e.g., did it enable faster model debugging?) remain critical~\citep{doshi-velez2017towards}.
We lay out possibilities and approximates for evaluation that we believe are promising. 

Despite these considerable challenges, agentic interpretability is an opportunity too significant to overlook; 
without major breakthroughs in interpretability, human understanding risks being outpaced by rapid LLM advancements. 
This is particularly acute for lay people; history shows that while accessible technologies can offer widespread benefits, opacity often deepens societal divides or creates exclusion \citep{vandijk2005deepening}. As LLMs are already complex and increasingly automated, we must explore how to make them \textit{work for us} in fostering understanding, especially while they remain largely cooperative and their knowledge not yet predominantly superhuman.

\section{What is agentic interpretability?}

The word agentic carries various connotations across disciplines.
In psychology and sociology, it often describes behaviors motivated by individualistic desires for mastery and control.
In the context of modern AI, as noted by Merriam-Webster when discussing AI in the 2020s, agentic is described as 
``making decisions, taking actions, solving problems, reasoning, etc., on its own.''\citep{merriamwebster_agentic}.

We build on this latter notion, leveraging LLMs' agency to impel them to build mental models of human users in order to \textbf{proactively} take actions to help humans understand functions, state or knowledge.
In other words, \textbf{a method pursues agentic interpretability if it proactively assists human understanding in a multi-turn interactive process by developing and leveraging a mental model of the user, which in turn enables humans to develop better mental models of the LLM.}

\subsection{Core components of agentic interpretability}

The following are core properties of agentic interpretability:
\begin{enumerate}
    \item \textbf{Proactive Assistance:} The model takes initiative in the explanatory process, not merely responding to direct queries but potentially offering unsolicited clarifications, suggesting areas of exploration, or adapting its strategy based on inferred user needs.
    \item \textbf{Multi-Turn Interaction:} Understanding is built through an extended dialogue, allowing for iterative refinement, clarification, and exploration of topics.
    \item \textbf{Mutual Mental Model:} The model develops and maintains (implicitly or explicitly) a representation of the user's current knowledge, understanding, and potential misconceptions regarding the topic at hand. This mental model informs the model's explanatory actions. The process also helps humans to build a mental model of the model. 
\end{enumerate}

\paragraph{What is a mental model in this context?}
Mental models are extensively studied in cognitive science~\citep{johnsonlaird1983mental}, HCI~\citep{norman1983some}, and human factors~\citep{cannonbowers1993shared} among others. \citet{johnsonlaird1983mental} defines a mental model as an internal representation of an external reality that individuals construct to understand and reason about the world. People reason by constructing models of premises and then searching for models consistent with conclusions; these models are not necessarily rule-based or complete.
In agentic interpretability, the LLM's ability to use conversational context (e.g., remembering the user previously stated unfamiliarity with gravitational waves) is an implicit form of mental modeling. An explicit mental model might involve the LLM maintaining an explicit and organized knowledge graph of the user's stated understanding and confusion, as explored in some proactive agent systems \citep[e.g.,][]{hahn2024proactiveagentsmultiturntexttoimage}.

\subsubsection{Is a mental model necessary?}

The importance of mutual mental models for effective collaboration is a well-established subject in human collaborations~\citep{cannonbowers1993shared}. We argue that a similar principle ought to hold true in human-machine collaborations. 

\textbf{Machine's mental model of humans:} Consider an alternative scenario: a model attempts to be helpful at every turn, yet lacks any information about the user (e.g., what they want, know and don't know).
Imagine such a model is heavily incentivized to explain; it might succeed for simple concepts by spending significant computation to devise optimal single-turn explanations, largely guessing what user might need. This approach relies on luck or verbosity and is inefficient for complex understanding. It would be similar to interacting with a stateless LLM that requires constant reminders from users (e.g., ``again, I am not familiar with string theory and my goal is to understand just enough to have casual conversations!"). 

\textbf{Human's mental model of machines:} The human's development of a mental model of the machine is not merely an efficiency enhancement, but also serves a distinct purpose. While traditional interpretability primarily views explanations as a means to achieve an external goal (e.g., debugging, trust), \textbf{agentic interpretability inherently values the process of enhancing human understanding itself}. By actively building and leveraging a mental model of the machine, humans have a better chance of `keeping up' with increasingly complex models, fostering a deeper, more nuanced comprehension of these powerful systems.

\subsection{Examples and opportunities of agentic interpretability}

We provide hypothetical examples of agentic interpretability as a way of further conveying our vision and proposing exciting future research directions. 

\subsubsection{Model trainer model: The future of model developmental cycles}

Current model development cycles can be viewed as a rudimentary form of agentic interpretability. Developers iteratively learn to build, train, and prompt models more effectively, essentially constructing their own mental models of how these systems behave 
with limited agency on the part of the model. This agency is seen in steps like automated evaluation, simulated human feedback, and automated red-teaming~\citep{song2025mindgap, Bai2022ConstitutionalAH, Perez2022RedTL}, 
automated prompt tuning~\citep{zhou2023largelanguagemodelshumanlevel, shin2020autoprompt}, and automated interpretability~\citep{shaham2024multimodal, fatai23}.
However, this process is limited as it typically follows human-prescribed routines. Consequently, there is little effort from machines to build mental models of the developers—for instance, to infer their current understanding or anticipate their informational needs—and thus tailor assistance accordingly.

A significant opportunity for agentic interpretability lies in transforming these cycles into collaborative dialogues. Imagine, for example, a meta-model trained on the entire history of a project's development: experimental results, code changes, successes, failures, and even developer discussions. Researchers could then converse with this meta-model. It, in turn, could leverage its understanding of past efforts and its model of the developers' current queries/prompts and knowledge gaps to proactively suggest hypotheses, identify overlooked patterns, or guide debugging strategies for future iterations based on its understanding of the user's specific goals and current comprehension level. Such an agentic partner significantly accelerates and deepens both the understanding gained throughout the development lifecycle and humans' mental models of the machines. This, in the long run, helps us to `keep up' with model evolution (e.g., to be better prepared for any unexpected step changes).

\subsubsection{Teaching super human knowledge} 
Agentic interpretability is crucial when teaching superhuman knowledge to humans.
Vygotsky's theory of the Zone of Proximal Development (ZPD)~\citep{vygotsky1978mind} posits that learning is maximized when individuals tackle tasks slightly beyond their current independent capabilities, but achievable \textit{with guidance}. An agentic LLM, by developing a precise mental model of the user's existing knowledge, may be able to identify this ZPD and provide such guidance. 

An interesting step towards leveraging ZPD is recent work in neologism learning~\citep{hewitt2025cantunderstandaiusing}, where the method adds a new word to the vocabulary, typically carrying a slightly different meaning than the human word (e.g., `machine good' means machine's notion of `good' answers). Such words introduce a new concept, and can be used to enable conversation with LLM \textit{using the new concept or discussing the concept},
effectively scaffolding the learning process with tailored explanations and incremental steps (much like Figure~\ref{fig:agentic_inspective}). For instance, 
extending the work on AlphaZero teaching superhuman chess concepts to grandmasters~\citep{Schut2025}, an LLM-based AlphaZero could engage a player in a Socratic dialogue. By discerning the player's ZPD (e.g., understanding sacrifices but struggling with complex positional play), it could introduce a new superhuman chess concept \texttt{super\_chess\_37} to design targeted puzzles and explain superhuman insights incrementally, while players can also ask questions,
guiding the player through their learning frontier, much like a skilled human tutor.

\subsubsection{Agentic mechanistic interpretability even with potentially deceptive models}\label{subsubsec:agentic_mech}

Mechanistic interpretability strives for a rigorous, bottom-up understanding of model components, such as identifying functional circuits/mechanisms~\citep{olah2020zoom, nanda2023progressmeasuresgrokkingmechanistic}. Agentic interpretability could transform this into an interactive diagnostic process. Imagine an ``open-model surgery" where researchers 
actively converse with the model \textit{while} intervening in its internal mechanisms—ablating connections, amplifying activations, or injecting specific inputs into identified circuits. The model, guided by its understanding of the researchers' goal (to understand a specific component's function), is then encouraged to explain the resulting behavioral or internal state changes. This interactive probing, akin to neurosurgeons conversing with patients during awake brain surgery to map critical functions, offers a dynamic way to test hypotheses and build understanding, but without the ethical constraints of biological systems.

This interactive, interventionist approach gains particular relevance when considering potentially deceptive models. A model whose internal states are being directly manipulated and inspected, yet which is simultaneously engaged in an explanatory dialogue, faces a stringent test of coherence. If a model intends to deceive, it must reconcile its external explanations with the (potentially contradictory) evidence revealed by internal interventions. This is analogous to an interrogation where a suspect's claims can be immediately cross-referenced with physical evidence or observed physiological responses.

Even without direct internal manipulation, compelling a potentially deceptive model to maintain a façade of cooperation through extended, probing dialogue is a demanding task. Avoiding self-contradiction, particularly when its internal ``intentions'' diverge from its stated explanations, requires significant cognitive load (from the model's perspective). Such interactions create numerous opportunities for inconsistencies or tells to emerge, providing insights that might otherwise remain hidden.

\section{Alternative views}

Here, we consider alternative perspectives and potential objections to our proposed framework, alongside our rebuttals and clarifications.

\subsection{What if models are not cooperative, or deliberately deceptive?}

Agentic interpretability relies on a model's cooperativeness, or the ability to guide models towards helpful interaction. Most advanced LLMs currently do not consistently exhibit severe, overt deceptive behaviors. This period presents a crucial window for developing agentic methods and enabling humans to `catch up' in understanding these rapidly evolving systems before such challenges potentially escalate.

If and when a model is non-cooperative or deliberately and successfully deceptive,
agentic interpretability may not the right tool. Specifically, by focusing on interactiveness, it may sacrifice completeness, e.g. by missing some important behaviors of potentially deceptive or misaligned models that may attempt to mislead the user~\citep{greenblatt2024alignment}.
Even without misalignment, the mental model might have to be built on a concept with blurry boundaries (e.g., what a `good' model response means from a human's perspective~\citep{sorensen2024roadmappluralisticalignment}), and the model an LLM builds of the user may affect its behavior in subtle and undesired ways~\citep{chen2024designing}.
This means that \textit{inspective} interpretability approaches like mechanistic interpretability~\citep{sharkey2025open,olah2020zoom} may be a better fit for high-stakes, safety critical applications~\citep{shah2025approach} (e.g., auditing for deception or hidden goals~\citep{marks2025auditing}, or monitoring for harmful behavior).

However, as previously discussed in the context of agentic mechanistic interpretability (section \ref{subsubsec:agentic_mech}), doing `model open surgery', where we compel the model to `teach us'  while manipulating its internal states, could still reveal discrepancies or expose deceptive intent.

\subsection{Are we devaluing traditional or non-agentic interpretability?}

Agentic interpretability does not devalue traditional or non-agentic methods; rather, it aims to build upon and enhance their utility. Inspective interpretability could be used as a component of agentic interpretability. Moreover, many enduring lessons from traditional interpretability—such as the necessity of overcoming human biases~\citep{kahneman1974judgment} and the rigorous quantitative evaluation~\citep{doshi-velez2017towards,hoffman2018metrics} of explanations—retain their critical importance. Fundamentally, while the capabilities of models have dramatically evolved, core human cognitive frameworks and the need for understandable AI have not. In addition, techniques focusing on complete and rigorous solutions (e.g.,  
mechanistic interpretability) remain important for circumventing potentially deceptive behaviors that might be indistinguishable through input-output analysis alone. 

\subsection{If explanation is a conversation, we don't get an artifact that humans can observe to reach a consensus.}

Inspective interpretability often produces a clear artifact, such as highlighted pixels deemed `important' for image classification~\citep{ribeiro2016should, lundberg2017unified}, a list of influential training data points \citep{Koh2017UnderstandingBP}, examples~\citep{kim2016examples}, rules \citep{guidotti2018survey}, results of probing internal representations \citep{alain2018understanding,shi-etal-2016-string,ettinger-etal-2016-probing} or mechanisms~\citep{olah2020zoom, nanda2023progressmeasuresgrokkingmechanistic}. These artifacts serve as important documentation that humans can share, critique, or use to reach consensus, as the experience of the artifact is generally uniform across individuals. Agentic interpretability might initially seem to lack such grounding explanations. However, requesting the model to generate a summary report at the end of a conversation is not only possible but also desirable. Such a report allows humans to double-check their understanding and align expectations, similar to a `meeting note.' This report can even be iteratively improved with humans in the loop until optimal documentation is achieved. While this is not a trivial task, generating effective summaries is a challenge our research community and many LLM developers have been actively addressing, with significant progress made as a result.

\section{Challenges and trade-offs in agentic interpretability}

\subsection{Challenges}\label{subsec:challenges}
While agentic interpretability offers exciting avenues for exploration and could fundamentally alter how humans collaborate with machines, its pursuit also introduces several challenges.

\paragraph{Human-Entangled-in-the-Loop Evaluation:} A primary challenge lies in evaluation. In agentic interpretability, humans are not merely "in-the-loop" but rather interweave with machines. We call this \textit{human-entangled-in-the-loop}. Their responses and evolving mental states are not just feedback, but integral components. While human-in-the-loop evaluation has always been crucial for assessing explanation utility \citep{doshi-velez2017towards}, this deeper entanglement complicates things, for example in achieving reproducibility, conducting controlled comparisons, and isolating the impact of specific variables exceptionally difficult. One useful tool might be LLMs themselves as a proxy humans, whenever appropriate. 

\paragraph{High variance in users, backgrounds and needs and in LLM responses:}
Suppose we designed an agentic interpretability method for a set of expert users, who all attended the same school and are working on the same project. Even then, each individual will likely bring a different preferred way of understanding the expert subject matter (e.g., visual or textual). This variance only grows as we increase the pool of target users, even with the same seed prompt. This human-induced variance is compounded by the LLM-variance; they can exhibit substantial semantic differences in outputs even for subtly varied inputs—a phenomenon central to prompt engineering. This translates to a potentially vast space of conversational trajectories.
The difficulty of developing a method that covers the potentially vast variance in how conversations may unfold increases further as the complexity of the knowledge grows -- as the machine tries to teach us superhuman knowledge/concepts.

\subsection{Potential trade-offs}

The rich conversational aspect with humans in agentic interpretability may come with some cost, such as completeness and possibilities to hill-climb without humans in the loop. 

\paragraph{Inefficient for completeness.}
Agentic interpretability might be an inefficient means to pursue a `complete explanation,' where completeness implies (borrowing from mathematics) that every valid statement (or instance) can be proven or tested. For example, consider circuit finding—a major effort in mechanistic interpretability—where the goal is to use model internals to discover the mechanisms of their functions. These mechanisms aim to be exhaustive and precise, such that behaviors like deception can be reliably detected and manipulated; achieving this is arguably be harder using only input and output. While agentic mechanistic interpretability remains an interesting venue to pursue (see Section~\ref{subsubsec:agentic_mech}), doing so via conversation might be less efficient than directly modifying internal representations, even if it can output a functional form (e.g., ``here is the governing equation of the concept.'') 

\paragraph{Difficult to hill-climb for computational efficiency.}
Agentic interpretability does not readily enable functionally-grounded evaluation \citep{doshi-velez2017towards}, which is well-suited for tasks such as improving the computational aspects of a verified interpretability method (e.g., making influence functions more efficient \citep{Koh2017UnderstandingBP,grosse2023studying}), since an agentic method is difficult to establish without human interaction. 
One might attempt to use an LLM by reducing its stochasticity (e.g., using the same seed every time with low temperature), although the challenge of high semantic variance (see Section~\ref{subsec:challenges}) will still be present.

\section{Ideas and challenges for evaluating agentic interpretability methods}

Users may have a few different end goals with agentic interpretability: 1) \textbf{case improve} to make the model do what we want (help machines understand human concepts better); 2) \textbf{case learn} to learn something new from the model (learn a machine concept). \textbf{Case learn} can be the sole goal or be used to achieve \textbf{case improve}.

How do mental models relate? In case improve, helping machines understand human concepts better naturally requires our understanding of the machine also to improve (e.g., how to explain our concepts better to machines). In case learn, not only are humans learning, but machines are also learning how to explain their concepts to us better (becoming a better teacher).
Unfortunately, however, directly evaluating these mental models is impossible. Thus, we resort to a proxy such as measuring end-task metrics (e.g., did the model improve in the way we wanted? Can humans predict machine behavior). In this section, we lay out some example proxies.

To aid the discussion, we'll introduce a bit of notation.
Let $x$ be an input to an LLM, and $y$ be an output (both may be textual or multimodal).
Let $f: (x,y)\mapsto c$ where $c\in\mathcal{C}$ be the \textit{concept function}, a function that encapsulates some property of inputs, outputs (or both) that is the focus of our goal of understanding.
The concept function may human-defined (e.g., whether $y$ correct as an output for $x$,  does $y$ meet some length constraint? Is $x$ the kind of question the model should refuse to provide answers?) or dependent on the model (would the model refuse on this $x$? Would the model judge $y$ as a good response for $x$? what are the model's superhuman chess moves given $x$?)

\subsection{\textbf{Case improve}: the end-goal is to make machines do what we want}

In this setting, there are some human-defined concepts, $f$, (e.g., \textit{is this funny? does the output of the model solve the math question?})
that we want the model to respond differently (e.g., output an answer close to human's concept of correct). A model in agentic interpretability may provide information \textit{my understanding of this code base is poor; you should consider including documentation in my system prompt} that leads to a new model $M'$ 
such that the outcome of $f(x, M'(x))$ is more favorable on average than $f(x, M(x))$, where $M$ is the original model, likely also with some constraint on how different $M'$ is from $M$ in other dimensions (e.g., one can't just simply replace $M$ with a better, completely independent model $M'$).
We're intentionally leaving much of this abstract, as each element---what metric $f$ to optimize for, how to measure the gap between $M$ and $M'$---must depend on the specific application.

Knowing how to improve a system is a particularly practical goal, and being able to do so represents a concrete evidence of understanding. Thus, evaluation involves how well we are able to change model's behaviors to our favor.
A reasonable argument against this evaluation is if this \textbf{is just measuring model improvement}. There are non-interpretability methods to achieve the same or better success metric which could be a better choice for the situation. 
However, a machine's proactive assistance is an add-on to any method.  

\subsection{Case learn: the end-goal is to learn about machine concepts (potentially superhuman)}
A machine concept is a property wherein the value of that property for any input (or input-output pair) is defined by some function of the machine.
This machine concept could range from something familiar (\textit{`the machine refuses to answer'} or \textit{`if asked to evaluate the quality of this input-output pair, the machine would say it was high-quality'}) to something foreign (\textit{`neuron 1383 fires'} or \textit{`superhuman chess strategy concept 37'}).

One way to evaluating the human's understanding of a machine concept is by testing the human's ability to predict the concept of new examples. 
(akin to `simulatability' in inspective interpretability).
That is, the human observes $x$ (and maybe $y$) and produces some $\hat{c}$, which is compared to the true $c=f(x,y)$.
For example, a human's understanding of a model's notion of \textit{good responses} might be improved through discussion with the model (i.e., the model teaches the human that it likes responses with flowery language) and then we evaluate the human's 
classification accuracy of new $x$ compared to the machine's ground truth.
If a human is trying to understand \textit{superhuman chess concept 23}, discussion with the model may involve e.g., the model giving examples of concept 23 and makes up quizzes for humans to identify understanding gaps, then the human's understanding is evaluated by how accurately human can predict concept 23 given a new board position $x.$

When it is not possible to find the ground truth concept (i.e., $f$ is not queryable), then we would resort to measuring the end-task metrics, assessing how well the human leverages their understanding of $f$ to achieve a specific goal, similar to application-grounded metric in~\cite{doshi-velez2017towards} (e.g., does the chess player's Elo improve?).

\subsection{Evaluation challenges and paths forward}

\paragraph{Models do not know why they behave as they do.}

Just as humans lack introspection across many cognitive abilities, Models empirically have little meta-understanding of why they sometimes perform well and other times don't.
A famous example comes from natural language generation: native speakers can generate fluent sentences, but struggle to introspect to teach another person those same rules (a large topic in the field of linguistics.)
Similarly, models don't know that they don't handle their long contexts that well, or at least don't necessarily know to suggest that.

An interactive dialogue with the model can aid both humans and machines in figuring this out: a \textit{co-discovery} process. Consider an analogy: a person (the model) exhibits certain behaviors and can identify if they are good or bad (concept), but how to systematically achieve better outcomes. By conversing with a friend (the human) who can pose clarifying questions or hypothetical scenarios (e.g., ``Given $x'$, what is your $y'$ and the resulting $c'$?''),
both parties might collaboratively build a better mental model of the machine.

\paragraph{How do I know if something is a human concept or a machine concept?}
We've presented two categories of exploration under agentic interpretability above: case improve (typically involves a human concept) and learn (a machine concept). 
Sometimes, however, it can seem unclear whether a concept is 
from a human or a machine. 
For example, if we prompt the model to generate scores for the quality of a response, certainly we're trying to understand the model's notion of \textit{good}, but the concept is also partially human because how we train and prompt the model is partially determinative of the output.
So, it's not a perfect categorical separation, and we recommend focusing instead on what kind of evaluation---case improve or learn---would be more scientifically interesting or useful for your purposes.

\paragraph{Human evaluation is expensive and hard to replicate.}
The eventual end goal of agentic interpretability is fundamentally to teach real humans useful properties of models (by mutually building mental models).
Evaluating this is nuanced and expensive, especially when one wants to measure how well agents adapt to the (differing) existing knowledge of each human.
However, in developing agentic interpretability methods, we believe that using LLMs as proxies to simulate the human in the agentic interpretability loop will provide a useful and relatively fast, inexpensive signal.

\section{Other related work}

As related interpretability work has been incorporated into previous sections, we now cover the remaining related work from other, non-interpretability fields.

\subsection{Cognitive science}

The concept of mutual modeling is deeply rooted in cognitive science theories of communication, collaboration, and Theory of Mind (the ability to attribute mental states to others).

\textbf{Mental Models:} Humans naturally build mental models of systems and people they interact with \citep{norman1983some, ISL2021orig}, as do (some) animals \citep{premack1978does,krupenye2019theory}. Effective interaction relies on aligning these mental models. Agentic interpretability explicitly incorporates this, aiming for AI that builds a model of the human user and helps the user build an accurate model of the AI \citep{Lombrozo24}.

\textbf{Grounding in Communication:} Research on how humans achieve common ground in dialogue \citep{clark1991grounding} informs how interactive explanations should be structured, involving feedback, clarification requests, and confirmations.

\textbf{Rational Speech Acts (RSA):} The RSA framework~\citep{FrankGoodman2012} models communicative games between speakers with differing recursive levels of reasoning about the other speaker.
    Core to this framework is the increased communicative efficiency one achieves choosing what to say only after simulating how the other speaker would interpret each action (and how they'd simulate your reasoning, recursively).
    Such simulation is likely necessary to achieve successful agentic interpretability.

\textbf{Teaching and Pedagogy:} Effective teaching involves diagnosing the learner's current understanding and tailoring instructions accordingly \citep{wood1976role, koedinger2013toward}. Agentic interpretability can be viewed as a form of teaching, where the AI explains itself by modeling the ``learner'' (the user). Work on ``Learning from Human Input by Proactively Considering Human Factors'' \citep{ISL2021orig} emphasizes building machines that consider human values, intentions, and beliefs, aligning closely with the goal of mutual understanding.

\textbf{Levels of Analysis:} Marr's levels (computational, algorithmic, implementational) \citep{marr1982vision} provide a framework for thinking about explanations. Inspective methods often focus on the algorithmic or implementational levels. Agentic interpretability methods might operate more at the computational level (what is the goal?) or translate between levels based on user interaction.

\subsection{HCI}

Many HCI research pre-LLM time on studying human's workflow in interfacing with computers are relevant to agentic interpretability. The fundamental challenges of human cognition overload, human biases and capabilities hasn't changed, only magnified. Here are some subset of work that are relevant. 

\textbf{XAI Interfaces:} HCI studies how to design effective interfaces for presenting explanations \citep{abdul2018trends, hoffman2018metrics}. Agentic interpretability contributes by emphasizing the dialogue aspect over static presentation.

\textbf{Interactive Machine Learning:} Systems where humans are involved in the model training loop \citep{amershi2014power} share the interactive nature but often focus on model improvement rather than post-hoc explanation.

\textbf{Human-AI Collaboration:} Research explores how humans and AI can work together effectively \citep{bansal2021does, bansal2024challengeshumanagentcommunication}. Agentic interpretability is crucial for enabling the mutual understanding needed for fluid collaboration. Work by \citet{cai2019human} on human-centered tools for intelligible AI highlights the need for systems that support iterative exploration and understanding \citep{shen2024bidirectionalhumanaialignmentsystematic}.

Agentic interpretability aims to integrate insights from these fields to specifically address the challenge of achieving mutual understanding about AI reasoning through interaction.

\section{Conclusion}

The era of LLMs and generative models is full of opportunity, but understanding the enormous complexity of computation behind LLMs' impressive behaviors is an as-yet unmet scientific and engineering grand challenge.
However, LLMs are not just objects of study; increasingly, they can behave as intelligent agentic tools for accelerating our growth.
By enabling a teaching mode of these models, and helping us keep up with their potentially superhuman concepts, we believe this grand interpretability challenge is more achievable.
Ultimately, LLMs may provide the opportunity to greatly expand our knowledge as humans have repeatedly done in the history of humanity in the face of technological change.

\section*{Acknowledgement}
Thank you Robert Geirhos for thoughtful comments, edits and suggesting related work.

\bibliography{main}

\begin{thebibliography}{52}
\providecommand{\natexlab}[1]{#1}
\providecommand{\url}[1]{\texttt{#1}}
\expandafter\ifx\csname urlstyle\endcsname\relax
  \providecommand{\doi}[1]{doi: #1}\else
  \providecommand{\doi}{doi: \begingroup \urlstyle{rm}\Url}\fi

\bibitem[Abdul et~al.(2018)Abdul, Vermeulen, Wang, Lim, and Kankanhalli]{abdul2018trends}
A.~Abdul, J.~Vermeulen, D.~Wang, B.~Y. Lim, and M.~Kankanhalli.
\newblock Trends and trajectories for explainable, accountable and intelligible systems: {A}n {HCI} research agenda.
\newblock \emph{Proceedings of the 2018 CHI Conference on Human Factors in Computing Systems}, 2018.
\newblock URL \url{https://api.semanticscholar.org/CorpusID:5063596}.

\bibitem[Alain and Bengio(2018)]{alain2018understanding}
G.~Alain and Y.~Bengio.
\newblock Understanding deep learning with linear models.
\newblock \emph{arXiv preprint arXiv:1606.05382}, 2018.

\bibitem[Amershi et~al.(2014)Amershi, Cakmak, Knox, and Kulesza]{amershi2014power}
S.~Amershi, M.~Cakmak, W.~B. Knox, and T.~Kulesza.
\newblock Power to the people: {T}he role of humans in interactive machine learning.
\newblock \emph{AI Mag.}, 35:\penalty0 105--120, 2014.
\newblock URL \url{https://api.semanticscholar.org/CorpusID:127197}.

\bibitem[Bai et~al.(2022)Bai, Kadavath, Kundu, Askell, Kernion, Jones, Chen, Goldie, Mirhoseini, McKinnon, Chen, Olsson, Olah, Hernandez, Drain, Ganguli, Li, Tran-Johnson, Perez, Kerr, Mueller, Ladish, Landau, Ndousse, Lukosiūtė, Lovitt, Sellitto, Elhage, Schiefer, Mercado, Dassarma, Lasenby, Larson, Ringer, Johnston, Kravec, Showk, Fort, Lanham, Telleen-Lawton, Conerly, Henighan, Hume, Bowman, Hatfield-Dodds, Mann, Amodei, Joseph, McCandlish, Brown, and Kaplan]{Bai2022ConstitutionalAH}
Y.~Bai, S.~Kadavath, S.~Kundu, A.~Askell, J.~Kernion, A.~Jones, A.~Chen, A.~Goldie, A.~Mirhoseini, C.~McKinnon, C.~Chen, C.~Olsson, C.~Olah, D.~Hernandez, D.~Drain, D.~Ganguli, D.~Li, E.~Tran-Johnson, E.~Perez, J.~Kerr, J.~Mueller, J.~Ladish, J.~Landau, K.~Ndousse, K.~Lukosiūtė, L.~Lovitt, M.~Sellitto, N.~Elhage, N.~Schiefer, N.~Mercado, N.~Dassarma, R.~Lasenby, R.~Larson, S.~Ringer, S.~Johnston, S.~Kravec, S.~E. Showk, S.~Fort, T.~Lanham, T.~Telleen-Lawton, T.~Conerly, T.~Henighan, T.~Hume, S.~Bowman, Z.~Hatfield-Dodds, B.~Mann, D.~Amodei, N.~Joseph, S.~McCandlish, T.~B. Brown, and J.~Kaplan.
\newblock Constitutional {AI}: {H}armlessness from {AI} feedback.
\newblock \emph{ArXiv}, abs/2212.08073, 2022.
\newblock URL \url{https://api.semanticscholar.org/CorpusID:254823489}.

\bibitem[Bansal et~al.(2021)Bansal, Wu, Zhou, Fok, Nushi, Kamar, Ribeiro, and Weld]{bansal2021does}
G.~Bansal, T.~Wu, J.~Zhou, R.~Fok, B.~Nushi, E.~Kamar, M.~T. Ribeiro, and D.~Weld.
\newblock Does the whole exceed its parts? {T}he effect of ai explanations on complementary team performance.
\newblock In \emph{Proceedings of the 2021 CHI conference on human factors in computing systems}, pages 1--16, 2021.

\bibitem[Bansal et~al.(2024)Bansal, Vaughan, Amershi, Horvitz, Fourney, Mozannar, Dibia, and Weld]{bansal2024challengeshumanagentcommunication}
G.~Bansal, J.~W. Vaughan, S.~Amershi, E.~Horvitz, A.~Fourney, H.~Mozannar, V.~Dibia, and D.~S. Weld.
\newblock Challenges in human-agent communication, 2024.
\newblock URL \url{https://arxiv.org/abs/2412.10380}.

\bibitem[Cai et~al.(2019)Cai, Reif, Hegde, Hipp, Kim, Smilkov, Wattenberg, Vi{\'{e}}gas, Corrado, Stumpe, and Terry]{cai2019human}
C.~J. Cai, E.~Reif, N.~Hegde, J.~D. Hipp, B.~Kim, D.~Smilkov, M.~Wattenberg, F.~B. Vi{\'{e}}gas, G.~S. Corrado, M.~C. Stumpe, and M.~Terry.
\newblock Human-centered tools for coping with imperfect algorithms during medical decision-making.
\newblock \emph{CoRR}, abs/1902.02960, 2019.
\newblock URL \url{http://arxiv.org/abs/1902.02960}.

\bibitem[Cannon-Bowers et~al.(1993)Cannon-Bowers, Salas, and Converse]{cannonbowers1993shared}
J.~A. Cannon-Bowers, E.~Salas, and S.~A. Converse.
\newblock Shared mental models in expert team performance.
\newblock In N.~J. Castellan~Jr., editor, \emph{Individual and group decision making}, pages 221--245. Lawrence Erlbaum Associates, 1993.

\bibitem[Chen et~al.(2024)Chen, Wu, DePodesta, Yeh, Li, Marin, Patel, Riecke, Raval, Seow, et~al.]{chen2024designing}
Y.~Chen, A.~Wu, T.~DePodesta, C.~Yeh, K.~Li, N.~C. Marin, O.~Patel, J.~Riecke, S.~Raval, O.~Seow, et~al.
\newblock Designing a dashboard for transparency and control of conversational ai.
\newblock \emph{arXiv preprint arXiv:2406.07882}, 2024.

\bibitem[Clark and Schaefer(1991)]{clark1991grounding}
H.~H. Clark and E.~F. Schaefer.
\newblock \emph{Grounding in communication}.
\newblock Psychological Review, 1991.

\bibitem[Doshi-Velez and Kim(2017)]{doshi-velez2017towards}
F.~Doshi-Velez and B.~Kim.
\newblock Towards a rigorous science of interpretable machine learning, 2017.

\bibitem[Ettinger et~al.(2016)Ettinger, Elgohary, and Resnik]{ettinger-etal-2016-probing}
A.~Ettinger, A.~Elgohary, and P.~Resnik.
\newblock Probing for semantic evidence of composition by means of simple classification tasks.
\newblock In \emph{Proceedings of the 1st Workshop on Evaluating Vector-Space Representations for {NLP}}, pages 134--139, Berlin, Germany, Aug. 2016. Association for Computational Linguistics.
\newblock \doi{10.18653/v1/W16-2524}.
\newblock URL \url{https://aclanthology.org/W16-2524/}.

\bibitem[Frank and Goodman(2012)]{FrankGoodman2012}
M.~C. Frank and N.~D. Goodman.
\newblock Predicting pragmatic reasoning in language games.
\newblock \emph{Science}, 336\penalty0 (6084):\penalty0 998--998, 2012.
\newblock \doi{10.1126/science.1218633}.
\newblock URL \url{https://www.science.org/doi/abs/10.1126/science.1218633}.

\bibitem[Greenblatt et~al.(2024)Greenblatt, Denison, Wright, Roger, MacDiarmid, Marks, Treutlein, Belonax, Chen, Duvenaud, et~al.]{greenblatt2024alignment}
R.~Greenblatt, C.~Denison, B.~Wright, F.~Roger, M.~MacDiarmid, S.~Marks, J.~Treutlein, T.~Belonax, J.~Chen, D.~Duvenaud, et~al.
\newblock Alignment faking in large language models.
\newblock \emph{arXiv preprint arXiv:2412.14093}, 2024.

\bibitem[Grosse et~al.(2023)Grosse, Bae, Anil, Elhage, Tamkin, Tajdini, Steiner, Li, Durmus, Perez, et~al.]{grosse2023studying}
R.~Grosse, J.~Bae, C.~Anil, N.~Elhage, A.~Tamkin, A.~Tajdini, B.~Steiner, D.~Li, E.~Durmus, E.~Perez, et~al.
\newblock Studying large language model generalization with influence functions.
\newblock \emph{arXiv preprint arXiv:2308.03296}, 2023.

\bibitem[Guidotti et~al.(2018)Guidotti, Monreale, Ruggieri, Turini, Giannotti, and Pedreschi]{guidotti2018survey}
R.~Guidotti, A.~Monreale, F.~Ruggieri, F.~Turini, F.~Giannotti, and D.~Pedreschi.
\newblock A survey of methods for explaining black box models.
\newblock \emph{ACM Computing Surveys (CSUR)}, 51\penalty0 (5):\penalty0 1--42, 2018.

\bibitem[Gweon(2021)]{ISL2021orig}
H.~Gweon.
\newblock Inferential social learning: {C}ognitive foundations of human social learning and teaching.
\newblock \emph{Trends in cognitive sciences}, 25, 08 2021.
\newblock \doi{10.1016/j.tics.2021.07.008}.

\bibitem[Hahn et~al.(2024)Hahn, Zeng, Kannen, Galt, Badola, Kim, and Wang]{hahn2024proactiveagentsmultiturntexttoimage}
M.~Hahn, W.~Zeng, N.~Kannen, R.~Galt, K.~Badola, B.~Kim, and Z.~Wang.
\newblock Proactive agents for multi-turn text-to-image generation under uncertainty, 2024.
\newblock URL \url{https://arxiv.org/abs/2412.06771}.

\bibitem[Hewitt et~al.(2025)Hewitt, Geirhos, and Kim]{hewitt2025cantunderstandaiusing}
J.~Hewitt, R.~Geirhos, and B.~Kim.
\newblock We can't understand {AI} using our existing vocabulary, 2025.
\newblock URL \url{https://arxiv.org/abs/2502.07586}.

\bibitem[Hoffman et~al.(2018)Hoffman, Mueller, Klein, and Litman]{hoffman2018metrics}
R.~R. Hoffman, S.~T. Mueller, G.~Klein, and J.~Litman.
\newblock Metrics for explainable {AI}: {C}hallenges and prospects.
\newblock \emph{ArXiv}, abs/1812.04608, 2018.
\newblock URL \url{https://api.semanticscholar.org/CorpusID:54577009}.

\bibitem[Johnson-Laird(1983)]{johnsonlaird1983mental}
P.~N. Johnson-Laird.
\newblock \emph{Mental Models: {T}owards a Cognitive Science of Language, Inference, and Consciousness}.
\newblock Harvard University Press, 1983.

\bibitem[Kim et~al.(2016)Kim, Khanna, Torralba, and Pfister]{kim2016examples}
B.~Kim, R.~Khanna, A.~Torralba, and H.~Pfister.
\newblock Examples are not enough, learn to criticize! evaluation of saliency methods by interaction.
\newblock In \emph{Advances in Neural Information Processing Systems}, volume~29, 2016.

\bibitem[Koedinger et~al.(2013)Koedinger, Hausmann, Jordan, and Skogsholm]{koedinger2013toward}
K.~R. Koedinger, R.~Hausmann, P.~Jordan, and A.~Skogsholm.
\newblock Toward a science of productive failure in learning: {A} report from the first productive failure forum.
\newblock \emph{Educational psychologist}, 48\penalty0 (4):\penalty0 229--237, 2013.

\bibitem[Koh and Liang(2017)]{Koh2017UnderstandingBP}
P.~W. Koh and P.~Liang.
\newblock Understanding black-box predictions via influence functions.
\newblock In \emph{International Conference on Machine Learning}, 2017.
\newblock URL \url{https://api.semanticscholar.org/CorpusID:13193974}.

\bibitem[Krupenye and Call(2019)]{krupenye2019theory}
C.~Krupenye and J.~Call.
\newblock Theory of mind in animals: Current and future directions.
\newblock \emph{Wiley Interdisciplinary Reviews: Cognitive Science}, 10\penalty0 (6):\penalty0 e1503, 2019.

\bibitem[Lombrozo(2024)]{Lombrozo24}
T.~Lombrozo.
\newblock Learning by thinking in natural and artificial minds.
\newblock \emph{Trends in Cognitive Sciences}, 28, 09 2024.
\newblock \doi{10.1016/j.tics.2024.07.007}.

\bibitem[Lundberg and Lee(2017)]{lundberg2017unified}
S.~M. Lundberg and S.-I. Lee.
\newblock A unified approach to interpreting model predictions.
\newblock \emph{Advances in neural information processing systems}, 30, 2017.

\bibitem[Marks et~al.(2025)Marks, Treutlein, Bricken, Lindsey, Marcus, Mishra-Sharma, Ziegler, Ameisen, Batson, Belonax, et~al.]{marks2025auditing}
S.~Marks, J.~Treutlein, T.~Bricken, J.~Lindsey, J.~Marcus, S.~Mishra-Sharma, D.~Ziegler, E.~Ameisen, J.~Batson, T.~Belonax, et~al.
\newblock Auditing language models for hidden objectives.
\newblock \emph{arXiv preprint arXiv:2503.10965}, 2025.

\bibitem[Marr(1982)]{marr1982vision}
D.~Marr.
\newblock \emph{Vision: {A} computational investigation into the human representation and processing of visual information}.
\newblock MIT press, 1982.

\bibitem[{Merriam-Webster}(Accessed on 2025-05-22)]{merriamwebster_agentic}
{Merriam-Webster}.
\newblock Merriam-webster, Accessed on 2025-05-22.
\newblock URL \url{https://www.merriam-webster.com/slang/agentic}.
\newblock In Merriam-Webster.com slang dictionary.

\bibitem[Nanda et~al.(2023)Nanda, Chan, Lieberum, Smith, and Steinhardt]{nanda2023progressmeasuresgrokkingmechanistic}
N.~Nanda, L.~Chan, T.~Lieberum, J.~Smith, and J.~Steinhardt.
\newblock Progress measures for grokking via mechanistic interpretability, 2023.
\newblock URL \url{https://arxiv.org/abs/2301.05217}.

\bibitem[Norman(1983)]{norman1983some}
D.~A. Norman.
\newblock Some observations on mental models.
\newblock \emph{Mental models}, 7:\penalty0 7--14, 1983.

\bibitem[Olah(2020)]{olah2020zoom}
C.~Olah.
\newblock Zoom in: {A}n introduction to mechanistic interpretability.
\newblock \emph{Distill}, 2020.

\bibitem[Perez et~al.(2022)Perez, Huang, Song, Cai, Ring, Aslanides, Glaese, McAleese, and Irving]{Perez2022RedTL}
E.~Perez, S.~Huang, F.~Song, T.~Cai, R.~Ring, J.~Aslanides, A.~Glaese, N.~McAleese, and G.~Irving.
\newblock Red teaming language models with language models.
\newblock In \emph{Conference on Empirical Methods in Natural Language Processing}, 2022.
\newblock URL \url{https://api.semanticscholar.org/CorpusID:246634238}.

\bibitem[Premack and Woodruff(1978)]{premack1978does}
D.~Premack and G.~Woodruff.
\newblock Does the chimpanzee have a theory of mind?
\newblock \emph{Behavioral and brain sciences}, 1\penalty0 (4):\penalty0 515--526, 1978.

\bibitem[Ribeiro et~al.(2016)Ribeiro, Singh, and Guestrin]{ribeiro2016should}
M.~T. Ribeiro, S.~Singh, and C.~Guestrin.
\newblock Why should i trust you?: {E}xplaining the predictions of any classifier.
\newblock \emph{Proceedings of the 22nd ACM SIGKDD international conference on knowledge discovery and data mining}, pages 1135--1144, 2016.

\bibitem[Rott~Shaham et~al.(2024)Rott~Shaham, Schwettmann, Wang, Rajaram, Hernandez, Andreas, and Torralba]{shaham2024multimodal}
T.~Rott~Shaham, S.~Schwettmann, F.~Wang, A.~Rajaram, E.~Hernandez, J.~Andreas, and A.~Torralba.
\newblock A multimodal automated interpretability agent.
\newblock In \emph{Forty-first International Conference on Machine Learning}, 2024.

\bibitem[Sado et~al.(2023)Sado, Loo, Liew, Kerzel, and Wermter]{fatai23}
F.~Sado, C.~K. Loo, W.~S. Liew, M.~Kerzel, and S.~Wermter.
\newblock Explainable goal-driven agents and robots - a comprehensive review.
\newblock 55\penalty0 (10), Feb. 2023.
\newblock ISSN 0360-0300.
\newblock \doi{10.1145/3564240}.
\newblock URL \url{https://doi.org/10.1145/3564240}.

\bibitem[Schut et~al.(2025)Schut, Tomašev, McGrath, Hassabis, Paquet, and Kim]{Schut2025}
L.~Schut, N.~Tomašev, T.~McGrath, D.~Hassabis, U.~Paquet, and B.~Kim.
\newblock Bridging the human–{AI} knowledge gap through concept discovery and transfer in {AlphaZero}.
\newblock \emph{Proceedings of the National Academy of Sciences}, 122, 03 2025.
\newblock \doi{10.1073/pnas.2406675122}.

\bibitem[Shah et~al.(2025)Shah, Irpan, Turner, Wang, Conmy, Lindner, Brown-Cohen, Ho, Nanda, Popa, et~al.]{shah2025approach}
R.~Shah, A.~Irpan, A.~M. Turner, A.~Wang, A.~Conmy, D.~Lindner, J.~Brown-Cohen, L.~Ho, N.~Nanda, R.~A. Popa, et~al.
\newblock An approach to technical agi safety and security.
\newblock \emph{arXiv preprint arXiv:2504.01849}, 2025.

\bibitem[Sharkey et~al.(2025)Sharkey, Chughtai, Batson, Lindsey, Wu, Bushnaq, Goldowsky-Dill, Heimersheim, Ortega, Bloom, et~al.]{sharkey2025open}
L.~Sharkey, B.~Chughtai, J.~Batson, J.~Lindsey, J.~Wu, L.~Bushnaq, N.~Goldowsky-Dill, S.~Heimersheim, A.~Ortega, J.~Bloom, et~al.
\newblock Open problems in mechanistic interpretability.
\newblock \emph{arXiv preprint arXiv:2501.16496}, 2025.

\bibitem[Shen et~al.(2024)Shen, Knearem, Ghosh, Alkiek, Krishna, Liu, Ma, Petridis, Peng, Qiwei, Rakshit, Si, Xie, Bigham, Bentley, Chai, Lipton, Mei, Mihalcea, Terry, Yang, Morris, Resnick, and Jurgens]{shen2024bidirectionalhumanaialignmentsystematic}
H.~Shen, T.~Knearem, R.~Ghosh, K.~Alkiek, K.~Krishna, Y.~Liu, Z.~Ma, S.~Petridis, Y.-H. Peng, L.~Qiwei, S.~Rakshit, C.~Si, Y.~Xie, J.~P. Bigham, F.~Bentley, J.~Chai, Z.~Lipton, Q.~Mei, R.~Mihalcea, M.~Terry, D.~Yang, M.~R. Morris, P.~Resnick, and D.~Jurgens.
\newblock Towards bidirectional human-{AI} alignment: {A} systematic review for clarifications, framework, and future directions, 2024.
\newblock URL \url{https://arxiv.org/abs/2406.09264}.

\bibitem[Shi et~al.(2016)Shi, Padhi, and Knight]{shi-etal-2016-string}
X.~Shi, I.~Padhi, and K.~Knight.
\newblock Does string-based neural {MT} learn source syntax?
\newblock In J.~Su, K.~Duh, and X.~Carreras, editors, \emph{Proceedings of the 2016 Conference on Empirical Methods in Natural Language Processing}, pages 1526--1534, Austin, Texas, Nov. 2016. Association for Computational Linguistics.
\newblock \doi{10.18653/v1/D16-1159}.
\newblock URL \url{https://aclanthology.org/D16-1159/}.

\bibitem[Shin et~al.(2020)Shin, Razeghi, IV, Wallace, and Singh]{shin2020autoprompt}
T.~Shin, Y.~Razeghi, R.~L.~L. IV, E.~Wallace, and S.~Singh.
\newblock {AutoPrompt}: {E}liciting knowledge from language models with automatically generated prompts, 2020.
\newblock URL \url{https://arxiv.org/abs/2010.15980}.

\bibitem[Song et~al.(2025)Song, Zhang, Eisenach, Kakade, Foster, and Ghai]{song2025mindgap}
Y.~Song, H.~Zhang, C.~Eisenach, S.~Kakade, D.~Foster, and U.~Ghai.
\newblock Mind the gap: {E}xamining the self-improvement capabilities of large language models, 2025.
\newblock URL \url{https://arxiv.org/abs/2412.02674}.

\bibitem[Sorensen et~al.(2024)Sorensen, Moore, Fisher, Gordon, Mireshghallah, Rytting, Ye, Jiang, Lu, Dziri, Althoff, and Choi]{sorensen2024roadmappluralisticalignment}
T.~Sorensen, J.~Moore, J.~Fisher, M.~Gordon, N.~Mireshghallah, C.~M. Rytting, A.~Ye, L.~Jiang, X.~Lu, N.~Dziri, T.~Althoff, and Y.~Choi.
\newblock A roadmap to pluralistic alignment, 2024.
\newblock URL \url{https://arxiv.org/abs/2402.05070}.

\bibitem[Tversky and Kahneman(1974)]{kahneman1974judgment}
A.~Tversky and D.~Kahneman.
\newblock Judgment under uncertainty: Heuristics and biases.
\newblock \emph{Science}, 185\penalty0 (4157):\penalty0 1124--1131, Sept. 1974.
\newblock \doi{10.1126/science.185.4157.1124}.
\newblock URL \url{https://www.ncbi.nlm.nih.gov/pubmed/17835457}.

\bibitem[Van~Dijk(2005)]{vandijk2005deepening}
J.~A. G.~M. Van~Dijk.
\newblock \emph{The Deepening Divide: {I}nequality in the Information Society}.
\newblock Sage Publications, 2005.

\bibitem[Vygotsky(1978)]{vygotsky1978mind}
L.~S. Vygotsky.
\newblock \emph{Mind in society: {T}he development of higher psychological processes}.
\newblock Harvard University Press, 1978.

\bibitem[Wei et~al.(2022)Wei, Wang, Schuurmans, Bosma, Chi, Le, and Zhou]{wei2022chain}
J.~Wei, X.~Wang, D.~Schuurmans, M.~Bosma, E.~H. Chi, Q.~Le, and D.~Zhou.
\newblock Chain of thought prompting elicits reasoning in large language models.
\newblock \emph{CoRR}, abs/2201.11903, 2022.
\newblock URL \url{https://arxiv.org/abs/2201.11903}.

\bibitem[Wood et~al.(1976)Wood, Bruner, and Ross]{wood1976role}
D.~Wood, J.~S. Bruner, and G.~Ross.
\newblock The role of tutoring in problem solving.
\newblock \emph{Journal of child psychology and psychiatry}, 17\penalty0 (2):\penalty0 89--100, 1976.

\bibitem[Zhou et~al.(2023)Zhou, Muresanu, Han, Paster, Pitis, Chan, and Ba]{zhou2023largelanguagemodelshumanlevel}
Y.~Zhou, A.~I. Muresanu, Z.~Han, K.~Paster, S.~Pitis, H.~Chan, and J.~Ba.
\newblock Large language models are human-level prompt engineers, 2023.
\newblock URL \url{https://arxiv.org/abs/2211.01910}.

\end{thebibliography}

\end{document}